\documentclass[10pt,twocolumn,letterpaper]{article}

\usepackage{iccv}
\usepackage{times}
\usepackage{epsfig}
\usepackage{graphicx}
\usepackage{amsmath}
\usepackage{amssymb}
\usepackage{color}
\usepackage{rotating}
\usepackage{afterpage}
\usepackage{subfig}

\usepackage[pagebackref=true,breaklinks=true,letterpaper=true,colorlinks,bookmarks=false]{hyperref}

\definecolor{john_color}{rgb}{0,0.5,1}
\definecolor{noah_color}{rgb}{0.2,0.64,0}
\definecolor{james_color}{rgb}{1,0.5,0}
\definecolor{tobias_color}{rgb}{0.5,1.0,0}

\newenvironment{packed_item}{
\begin{itemize}
  \setlength{\itemsep}{1pt}
  \setlength{\parskip}{0pt}
  \setlength{\parsep}{0pt}
}{\end{itemize}}

\newcommand{\leaveout}[1]{}
\iccvfinalcopy % *** Uncomment this line for the final submission

\def\iccvPaperID{****} % *** Enter the ICCV Paper ID here

\ificcvfinal\fi
\begin{document}

\title{DeepStereo: Learning to Predict New Views from the World's Imagery}

\author{John Flynn\\
Google Inc.\\
{\tt\small jflynn@google.com}
\and
Ivan Neulander\\
Google Inc.\\
{\tt\small ineula@google.com}
\and
James Philbin\\
Google Inc.\\
{\tt\small jphilbin@google.com}
\and
Noah Snavely\\
Google Inc.\\
{\tt\small snavely@google.com}
}

\maketitle

\begin{abstract}
Deep networks have recently enjoyed enormous success when applied to recognition
and classification problems in computer vision~\cite{krizhevsky12,szegedy14},
but their use in graphics problems has been limited
(\cite{kulkarni15,dosovitskiy15learning} are notable recent exceptions). In this
work, we present a novel deep architecture that performs new view synthesis
directly from pixels, trained from a large number of posed image sets. In contrast to
traditional approaches which consist of multiple complex stages of processing,
each of which require careful tuning and can fail in unexpected ways, our system
is trained end-to-end. The pixels from neighboring views of a scene are
presented to the network which then directly produces the pixels of the unseen
view. The benefits of our approach include generality (we only require posed
image sets and can easily apply our method to different domains), and high
quality results on traditionally difficult scenes.  We believe this is due to
the end-to-end nature of our system which is able to plausibly generate pixels
according to color, depth, and texture priors learnt automatically from the
training data. To verify our method we show that it can convincingly reproduce 
known test views from nearby imagery. Additionally we show images rendered from 
novel viewpoints.
To our knowledge, our work is the first to apply deep learning to the problem of 
new view synthesis from sets of real-world, natural imagery.

\end{abstract}

\section{Introduction}

\begin{figure}%[h!]
\centering
\subfloat{
  \includegraphics[width=\linewidth]{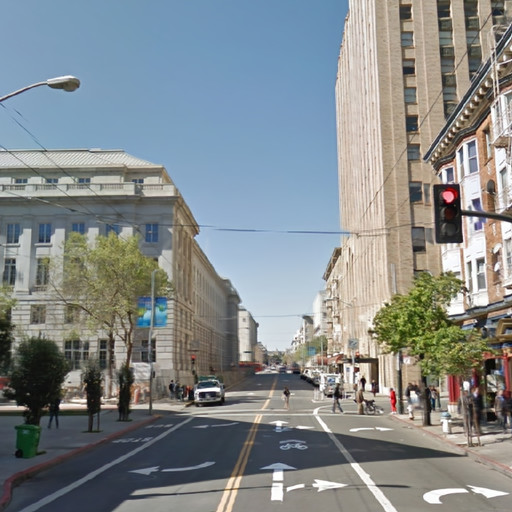}
}\\
\vspace{-.125in}
\subfloat{
  \includegraphics[width=.5\columnwidth]{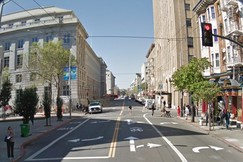}
  \includegraphics[width=.5\columnwidth]{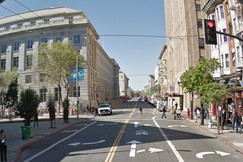}
}
\caption{The top image was synthesized from several input panoramas. A portion of two of the inputs is shown on the bottom row. \\ More results at: \url{http://youtu.be/cizgVZ8rjKA}}
\label{fig:front_page_result}
\end{figure}

Estimating 3D shape from multiple posed images is a fundamental task
in computer vision and graphics, both as an aid to image understanding
and as a way to generate 3D representations of scenes that can be
rendered and edited. In this work, we aim to solve the related problem
of \textit{new view synthesis}, a form of image-based rendering (IBR) where the goal is to synthesize a new
view of a scene by warping and combining images from nearby posed
images. This can be used for applications such as cinematography,
virtual reality, teleconferencing~\cite{criminisi07}, image
stabilization~\cite{kopf14}, or 3-dimensionalizing monocular film
footage.

New view synthesis is an extremely challenging, under-constrained problem.  An
exact solution would require full 3D knowledge of all visible geometry in the
unseen view which is in general not available due to occluders. Additionally,
visible surfaces may have ambiguous geometry due to a lack of
texture. Therefore, good approaches to IBR typically
require the use of strong priors to fill in pixels where the geometry is
uncertain, or when the target color is unknown due to occlusions.

The majority of existing techniques for this problem involve
traditional multi-view stereo and/or image warping methods and often
explicitly model the stereo, color, and occlusion components of each
target pixel~\cite{woodford07,chaurasia13depth}. A key problem with these approaches
is that they are prone to generating unrealistic and jarring
rendering artifacts in the new view. Commonly seen artifacts include
tearing around occluders, elimination of fine structures, and
aliasing. Handling complex, self-occluding (but commonly seen) objects
such as trees is particularly challenging for traditional approaches.
Interpolating between wide baseline views tends to exacerbate these
problems.

Deep networks have enjoyed huge success in recent years, particularly for image
understanding tasks~\cite{krizhevsky12,szegedy14}.  Despite these successes,
relatively little work exists on applying deep learning to computer graphics problems and
especially to generating new views from real imagery. One possible reason is the
perceived inability of deep networks to generate pixels directly, but recent
work on denoising~\cite{xie12}, super-resolution~\cite{dong14}, and
rendering~\cite{kulkarni15} suggest that this is a misconception. Another common
objection is that deep networks have a huge number of parameters and hence are
prone to overfitting in the absence of enormous quantities of data, but recent
work~\cite{szegedy14} has demonstrated state-of-the-art deep networks whose
parameters number in the low millions, greatly reducing the potential for
overfitting.

In this work we present a new approach to new view synthesis
that uses deep networks to regress directly to output pixel colors
given the posed input images. Our system is able to interpolate between views 
separated by a wide baseline and exhibits resilience to traditional failure 
modes, including graceful degradation in the presence of scene motion and 
specularities. We posit
this is due to the end-to-end nature of the training, and the ability
of deep networks to learn extremely complex non-linear functions of
their inputs~\cite{montufar14}. Our method makes minimal assumptions
about the scene being rendered: largely, that the scene should be
static and should exist within a finite range of depths. Even when
these requirements are violated, the resulting images degrade
gracefully and often remain visually plausible.
When uncertainty cannot be avoided our method prefers to blur detail
which generates much more visually pleasing results compared to
tearing or repeating, especially when animated.
Additionally, although we focus on its application to new view
problems here, we believe that the deep architecture presented can be
readily applied to other stereo and graphics problems given suitable
training data.

For view synthesis, there is an abundance of readily available
training data---any set of posed images can be used as a training
set by leaving one image out and trying to reproduce it from the
remaining images. We take that approach here, and train our models
using large amounts of data mined from Google's Street View, a massive
collection of posed imagery spanning much of the
globe~\cite{klinger_2013}. Because of the variety of the scenes seen
in training our system is robust and generalizes to indoor and outdoor
imagery, as well as to image collections used in prior work.

We compare images generated by our model with the corresponding captured
images on street and indoor scenes. Additionally, we compare our results 
qualitatively to existing state-of-the-art IBR methods.

\section{Related Work}\label{sec:related}

\noindent{\bf Learning depth from images.} The problem of view synthesis is
strongly related to the problem of predicting depth or 3D shape from imagery. In recent
years, learning methods have been applied to this shape prediction problem,
often from just a single image---a very challenging vision task. Automatic
single-view methods include the Make3D system of Saxena
\etal~\cite{saxena09make3d}, which uses aligned photos and laser scans as
training data, and the automatic photo pop-up work of Hoiem
\etal~\cite{hoiem05popup}, which uses images with manually annotated geometric
classes. More recent methods have used Kinect data for
training~\cite{karsch14depth,konrad12conversion} and deep learning methods for
single view depth or surface normal
prediction~\cite{eigen14depth,wang14designing}. However, the single-view problem
remains very challenging. Moreover, gathering sufficient training data is
difficult and time-consuming.

Other work has explored the use of machine learning for the stereo problem
(i.e., using more than one frame). Learning has been used in several ways, including
estimating the parameters of more traditional models such as
MRFs~\cite{yamaguchi12continuous} and learning low-level correlation filters for
disparity estimation~\cite{memisevic11stereopsis,konda13unsupervised}.

Unlike
this prior work, we learn to synthesize new views directly using a new deep
architecture, and do not require known depth or disparity as training data.

\medskip
\noindent{\bf View interpolation.} There is a long history of
work on image-based rendering in vision and graphics based on a variety of
methods, including light fields~\cite{levoy96lightfield,gortler96lumigraph},
image correspondence and warping~\cite{seitz96view}, and explicit shape and
appearance
estimation~\cite{vedula05sceneflow,zitnick04highquality,shan13visual}.  Much of
the recent work in this area has used a combination of 3D shape with image
warping and
blending~\cite{eisemann08floating,goesele10ambient,chaurasia13depth,chaurasia11silhouette}. These
methods are largely hand-built and do not leverage training data. Our goal is to
learn a model for predicting new viewpoints by directly minimizing the prediction error on
our training set.

We are particularly inspired by the work of Fitzgibbon \etal on IBR using
image-based priors~\cite{fitzgibbon03ibr}. Like them, we consider the
goal of faithfully reconstructing the actual output image to be the
key problem to be optimized for, as opposed to reconstructing depth
or other intermediate representations. We utilize state-of-the-art
machine learning methods with a new architecture to achieve this goal.
Szeliski \cite{szeliski99prediction} suggests using image 
prediction error as a metric for stereo algorithms; our method directly 
minimizes this prediction error.

Finally, a few recent papers have applied deep learning to
synthesizing imagery. Dosovitskiy \etal train a network on synthetic
images of rendered 3D chairs that can generate new chair images given
parameters such as pose~\cite{dosovitskiy15learning}. Kulkarni \etal
propose a ``deep convolutional inverse graphics network'' that can
parse and rerender imagery such as faces~\cite{kulkarni15deep}.
However, we believe ours is the first method to apply deep learning to
synthesizing novel natural imagery from posed real-world input images.

\section{Approach}\label{sec:approach}

\begin{figure}[h!]
\centering \includegraphics[width=\columnwidth]{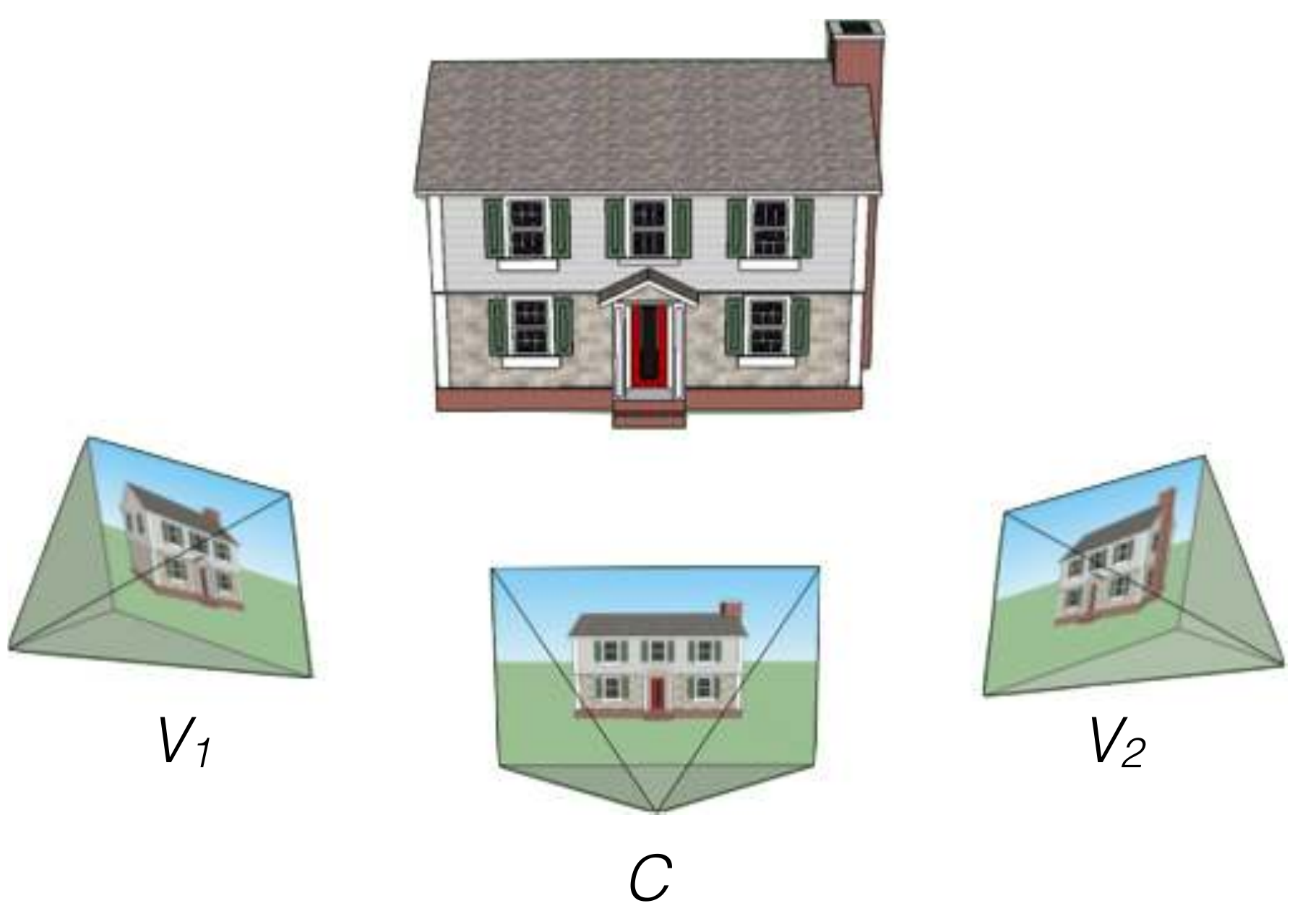}
\caption{The goal of image-based rendering is to render a new view at $C$ from
  existing images at $V_1$ and $V_2$.}
\label{fig:image_based_rendering}
\end{figure}
Given a set of posed input images $I_1, I_2, \dots, I_n$, with poses $V_1,
  V_2, \dots, V_n$, the view synthesis problem is to render a new image from
the viewpoint of a new target camera $C$
(Fig.~\ref{fig:image_based_rendering}). Despite the representative power of deep
networks, naively training a deep network to synthesize new views by supplying
the input images $I_k$ as inputs directly is unlikely to work well, for two key
reasons.

First, the pose parameters of $C$ and of the views $V_1, V_2, \dots, V_n$ would 
need to be supplied as inputs to the network in order to produce the desired 
view. The relationship between the pose parameters, the input pixels and the 
output pixels is complex and non-linear---the network would effectively need to 
learn how to interpret rotation angles and to perform image reprojection.
Requiring the network to learn projection is inefficient---it is a
straightforward operation that we can represent outside of the network.

Second, in order to synthesize a new view, the network would need to compare and
combine potentially distant pixels in the original source images, necessitating
very dense, long-range connections. Such a network would have many parameters and would be
slow to train, prone to overfitting, and slow to run inference on. It is
possible that a network structure could be designed to use the epipolar
constraint internally in order to limit connections to those on corresponding
epipolar lines. However, the epipolar lines, and thus the network connections,
would be pose-dependent, making this very difficult and likely computationally
inefficient in practice. 

\begin{figure}[h!]
  \centering \includegraphics[width=\columnwidth]{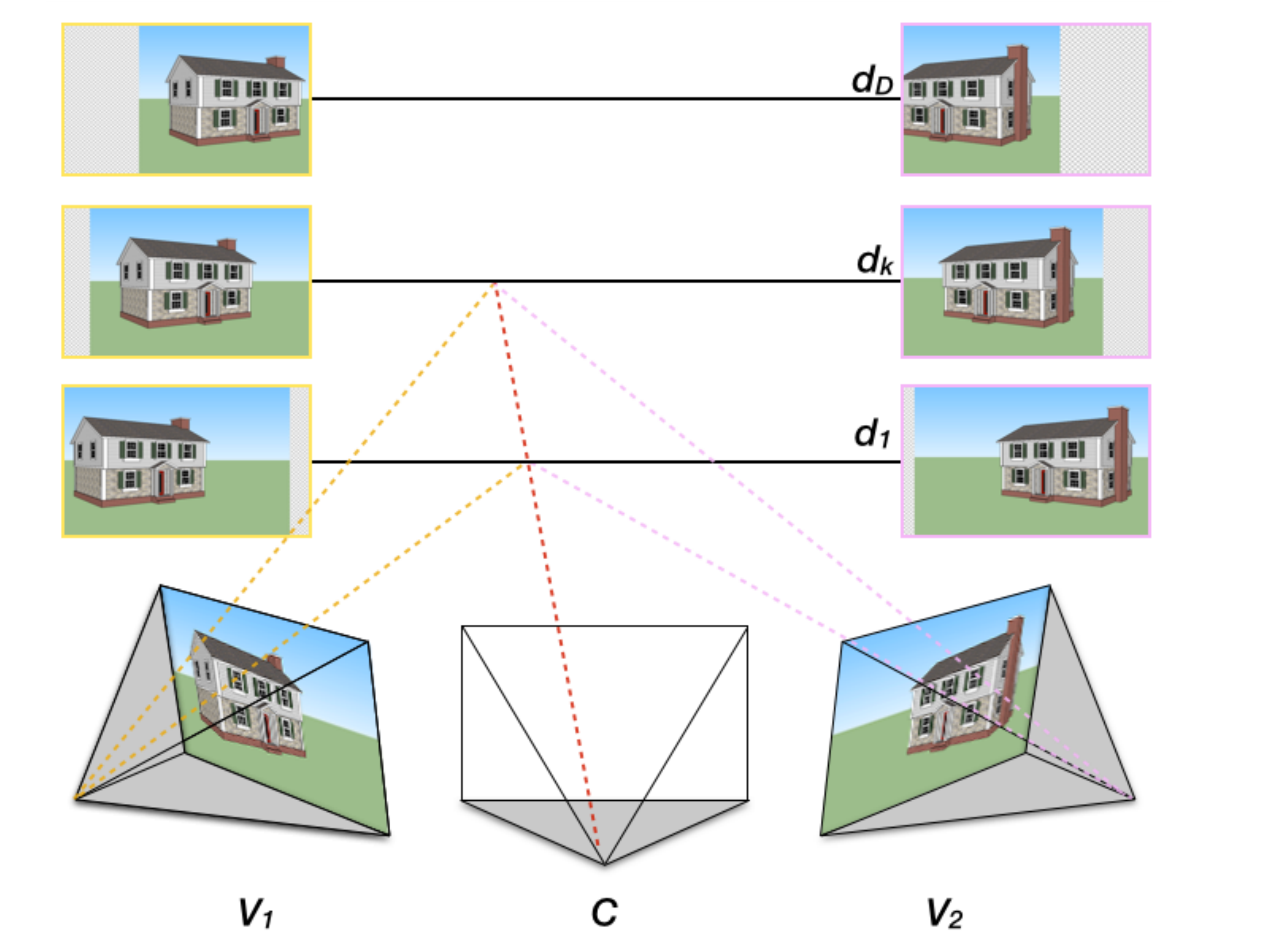}
  \caption{Plane sweep stereo reprojects images $I_1$ and $I_2$ from viewpoints $V_1$ and $V_2$
    to the target camera $C$ at a range of depths $d \in {d_1 \dots d_D}$. The dotted rays indicate the pixels from the input images reprojected to a particular output image pixel, and the images above each input view show the corresponding reprojected images at different depths.}
  \label{fig:plane_sweep_stereo}
\end{figure}

\medskip
\noindent{\bf Using plane-sweep volumes.} Instead, we address these problems by using ideas from traditional
plane sweep stereo \cite{collins96spacesweep,
  szeliski99stereomatching}. We provide our
network with a set of 3D plane sweep volumes as input. A plane
sweep volume consists of a stack of images reprojected to the target
camera $C$ (Fig.~\ref{fig:plane_sweep_stereo}).  Each image $I_k$ in the stack is reprojected into the target
camera $C$ at a set of varying depths $d \in \{d_1, d_2, \dots d_D\}$ to form a
plane sweep volume $V^k_C = \{P^k_1, P^k_2, \dots P^k_D\}$, where $P^k_i$ refers to the reprojected image $I_k$ at depth $d_i$. 
Reprojecting an input image into the target camera only requires basic texture mapping
capabilities and can be performed on a GPU.  A separate plane sweep
volume $V^k_C$ is created for each input image $I_k$. Each voxel
$v^k_{i,j,z}$ in each plane sweep volume $V^k_C$ has $R$, $G$, $B$ and $A$ (alpha)
components. The alpha channel indicates the availability of source pixels for 
that voxel (e.g., alpha is zero for pixels outside the field of view of a source image).

Using plane sweep volumes as input to the network removes the need to supply the
pose parameters since they are now implicit inputs used in the construction of
the plane sweep volume. Additionally, the epipolar constraint is trivially
enforced within a plane sweep volume: corresponding pixels are now in
corresponding $i,j$ columns of the plane sweep volumes. Thus, long range connections between pixels are no longer
needed, so a given output pixel depends only on a small column of voxels from
each of the per-source plane sweep volumes. Similarly, the computation performed
to produce an output pixel $p$ at location $i,j$ should be largely independent of 
the pixel location. This allows us to use more efficient convolutional neural 
networks. Our model applies 2D convolutional layers to each plane within the input plane 
sweep volume.
In addition to sharing weights within convolutional layers we make
extensive use of weight sharing across planes in the plane sweep
volume. Intuitively, weight sharing across planes make sense since the 
computation to be performed on each plane will be largely independent of the 
plane's depth.

\begin{figure}[h!]
  \centering \includegraphics[width=\columnwidth]{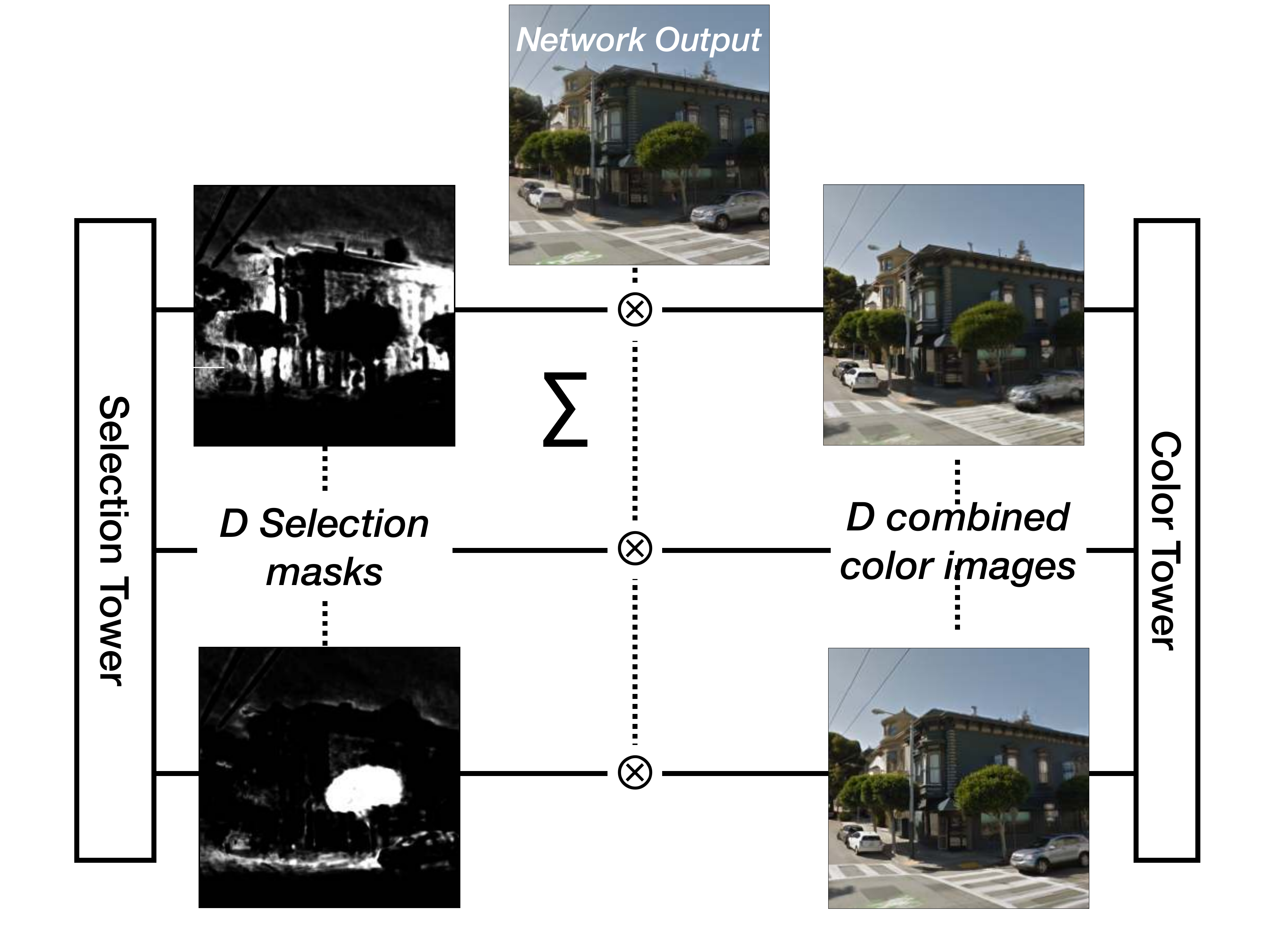}
  \caption{The basic architecture of our network, with selection and color
    towers. The final output image is produced by element-wise multiplication of
    the selection and color tower outputs and then computing the sum over the
    depth planes.  Fig.~\ref{fig:full_network} shows the full complete network
    details.}
  \label{fig:output_combination}
\end{figure}

\medskip
\noindent{\bf Our model.}  Our network architecture
(Fig.~\ref{fig:output_combination}) consists of two towers of layers, a {\em
  selection tower} and a {\em color tower}. The intuition behind this dual
 network architecture is that there are there are really two related tasks 
that we are trying to accomplish:
\begin{packed_item}
\item{\bf Depth prediction.} First, we want to know the approximate depth for
  each pixel in the output image. This enables us to determine the source image
  pixels we should use to generate that output pixel. In prior work, this kind
  of probability over depth might be computed via SSD, NCC, or variance; we
  learn how to compute these probabilities using training data.
\item{\bf Color prediction.} Second, we want to produce a color for that output
  pixel, given all of the relevant source image pixels. Again, the network does
  not just perform, e.g., a simple average, but learns how to optimally combine
  the source image pixels from training data.
\end{packed_item}
The two towers in our network correspond to these two tasks: the selection tower
produces a probability map (or ``selection map'') for each depth indicating the
likelihood of each pixel having a given depth. The color tower produces a full
color output image for each depth; one can think of this tower as producing the
best color it can for each depth, assuming that the depth is the correct
one. These $D$ color images are then combined by computing a per-pixel weighted
sum with weights drawn from the selection maps---the selection maps decide on
the best color layers to use for each output pixel. This simple new approach to
view synthesis has several attractive properties. For instance, we can learn all
of the parameters of both towers simultaneously, end-to-end using deep learning
methods. The weighted averaging across color layers also yields some resilience
to uncertainty---regions where the algorithm is not confident tend to be blurred
out, rather than being filled with warped or distorted input pixels.

More formally, the selection tower computes, for each pixel $p_{i,j}$, in each
plane $P_z$, the \textit{selection probability} $s_{i,j,z}$ for the pixel being
at that depth.  The color tower computes for each pixel $p_{i,j}$ in each plane
$P_z$ the color $c_{i,j,z}$ for the pixel at that plane. The final output color
for each pixel is computed as a weighted summation over the output color planes,
weighted by the selection probability (Fig.~\ref{fig:output_combination}):
%%%
\begin{equation}
c^f_{i,j} = \sum s_{i,j,z} \times c_{i,j,z}.
\label{eq:combine}
\end{equation}
%%%

The input to each tower is the set of plane sweep volumes $V^k_C$. The first
layer of both towers concatenates the input plane sweep volumes over the
source. This allows the networks to compare and combine reprojected
pixel values across sources. We now describe the computation performed in each 
tower in more detail.

\begin{figure}[t]
  \centering \includegraphics[width=\columnwidth]{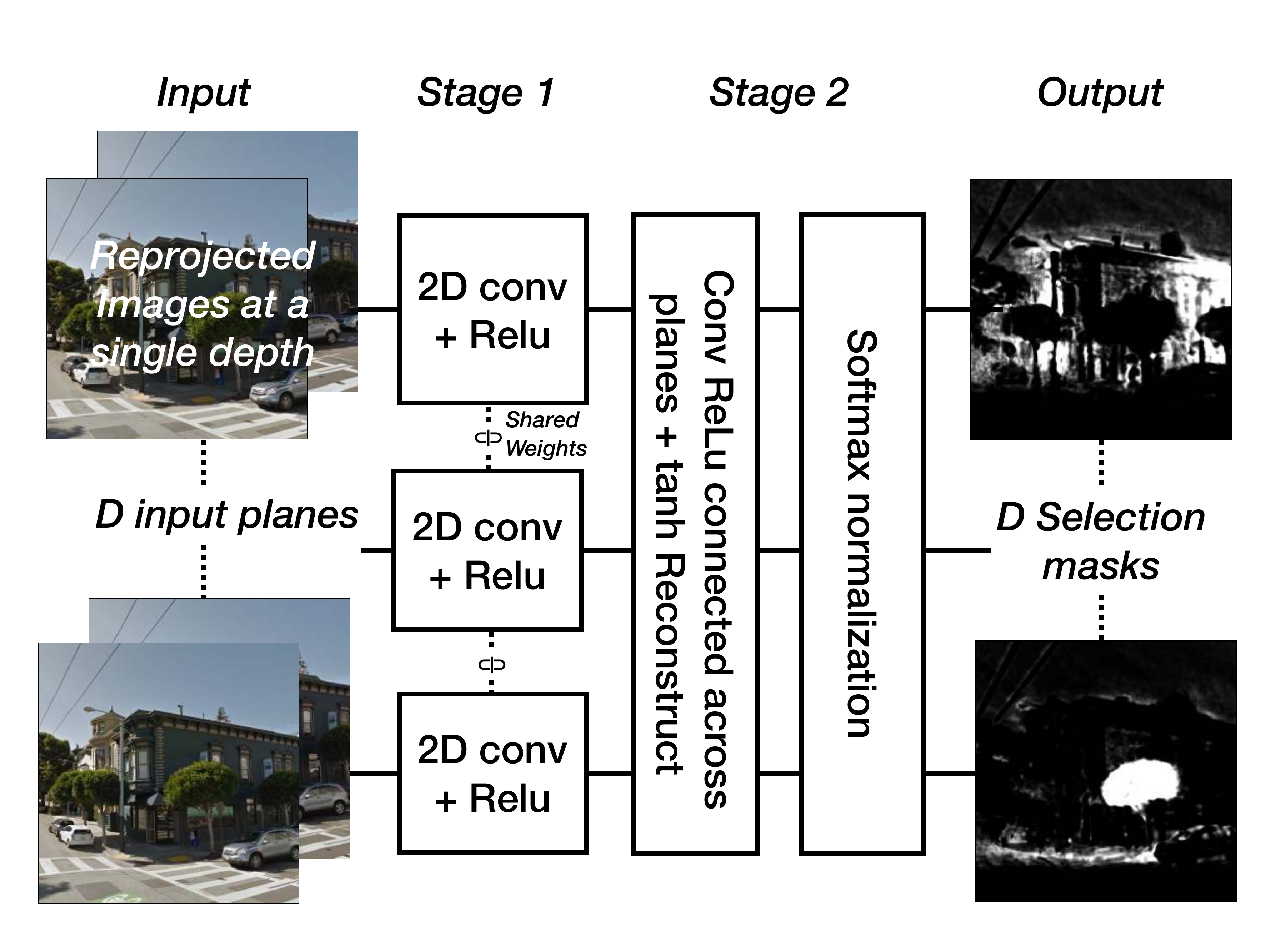}
  \caption{The selection tower learns to produce a selection probability
    $s_{i,j,z}$ for each pixel $p_{i,j}$ in each depth plane $P_z$.}
  \label{fig:selection_tower}
\end{figure}
%%%
\medskip
\noindent{\bf The selection tower.}  The selection tower consists of two main
stages. The first stage is a number of $2D$ convolutional rectified linear
layers that share weights across all planes. Intuitively the early layers will
compute features that are independent of depth, such as pixel differences, so
their weights can be shared. The second stage of layers are connected across
depth planes, in order to model interactions between depth planes such as those
caused by occlusion (e.g., the network might learn to prefer closer planes that
have high scores in case of ambiguities in depth). The final layer of the
network is a per-pixel \textit{softmax} normalization transformer over
depth. The \textit{softmax} transformer encourages the model to pick a single
depth plane per pixel, whilst ensuring that the sum over all depth planes is
1. We found that using a $\tanh$ activation for the penultimate layer gives more
stable training than the more natural choice of a linear layer. In our
experiments the linear layer would often ``shut down'' certain depth
planes\footnote{The depth planes would receive zero weight for all inputs and
  all pixels.} and never recover, presumably, due to large gradients from the
\textit{softmax} layer. The output of the selection tower is a 3D volume of
single-channel nodes $s_{i,j,z}$ where
$$\sum_{z=1}^{D} s_{i,j,z} = 1.$$

\begin{figure}[t]
  \centering
    \includegraphics[width=\columnwidth]{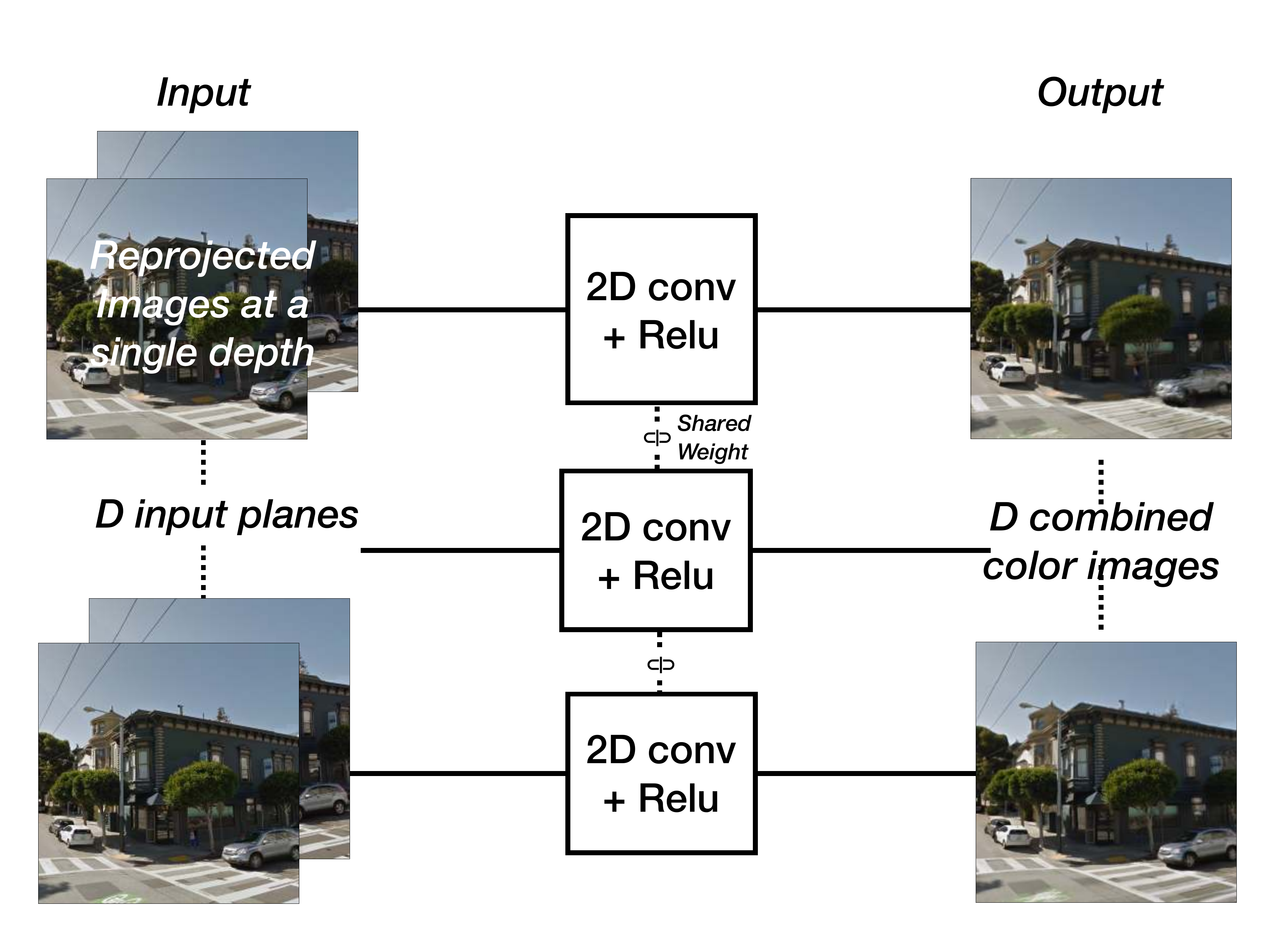}
  \caption{The color tower learns to combine and warp pixels across sources to
    produce a color $c_{i,j,z}$ for each pixel $p_{i,j}$ in each depth plane
    $P_z$.}
  \label{fig:color_tower}
\end{figure}

\noindent{\bf The color tower.} The color tower (Fig.~\ref{fig:color_tower}) is
simpler and consists of only 2D convolutional rectified linear layers that share
weights across all planes, followed by a linear reconstruction layer. Occlusion
effects are not relevant for the color layer so no across-depth interaction is
needed. The output of the color tower is again a 3D volume of nodes
$c_{i,j,z}$. Each node in the output has 3 channels, corresponding to $R$, $G$
and $B$.

The output of the color tower and the selection tower are multiplied together
per node to produce the output image $c^f$ (Eq.~\ref{eq:combine}). During training the resulting image is
compared with the known target image $I^t$ using a per-pixel $L_1$ loss. The
total loss is thus: 
%%%%
$$L = \sum_{i,j} |c^t_{i,j} - c^f_{i,j}|$$
%%%%
where $c^t_{i,j}$ is the target color at pixel $i,j$.

\medskip
\noindent {\bf Multi-resolution patches.}  Rather than predict a full image at a
time, we predict the output image patch-by-patch. We found that passing in a set
of lower resolution versions of successively larger areas around the input
patches helped improve results by providing the network with more context. We
pass in four different resolutions. Each resolution is first processed
independently by several layers and then upsampled and concatenated before
entering the final layers. The upsampling is performed using nearest neighbor
interpolation.

\begin{figure*}[t]
  \centering
    \includegraphics[width=\textwidth]{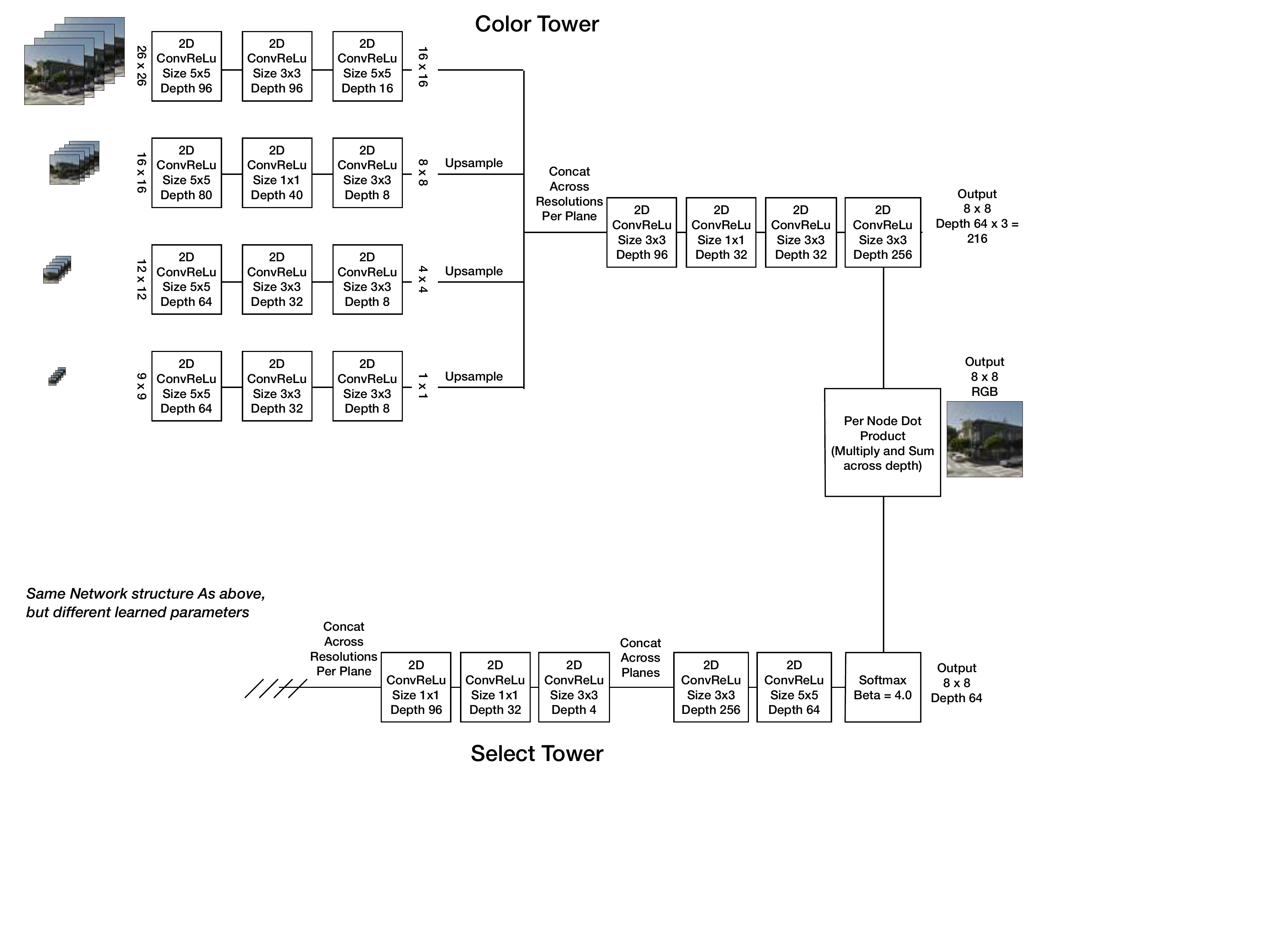}
    \caption{The full network diagram. The initial stages of both the color 
             and select towers are the same structure, but do not share parameters.} 
    \label{fig:full_network}
\end{figure*}

The full details of the complete network are shown in Fig.~\ref{fig:full_network}.

\subsection{Training}
To train our network, we used images of street scenes captured by a moving
vehicle. The images were posed using a combination of odometry and traditional
structure-from-motion techniques \cite{klinger_2013}. The vehicle captures a set
of images, known as a rosette, from different directions for each exposure. The
capturing camera uses a rolling shutter sensor, which is taken into account by
our camera model.  We used approximately 100K of such image sets during
training.

We used a continuously running online sample generation pipeline that selected
and reprojected random patches from the training imagery. The network was
trained to produce $8\times8$ patches from overlapping input patches of size
$26\times26$. We used 96 depth planes in all results shown. Since the network is 
fully convolutional there are no border effects as we transition between patches in the output image. In order to
increase the variability of the patches that the network sees during training
patches from many images are mixed together to create mini-batches of size
400. We trained our network with Adagrad \cite{adagrad} with an initial learning
rate of 0.0005 using the system described by Dean, \etal~\cite{brain}.  In our
experiments, training converged after approximately 1M steps.  Due to sample
randomization, it is unlikely that any patch was used more than once in
training. Thanks to our large volume of training data, training data
augmentation was not required.  We selected our training data by first randomly
selecting two rosettes that were captured relatively close together, within 30cm, we
then found other nearby rosettes that were spaced up to 3m away. We select one of the images in the center rosette as the
target and train to produce it from the others. 

\section{Results}\label{sec:results}
\begin{figure}%
\centering
\subfloat[Our result.]{
  \includegraphics[width=.5\linewidth]{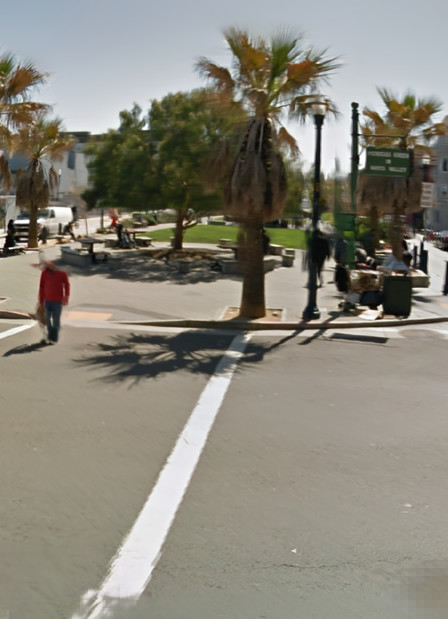}
}
\subfloat[Reference image.]{
  \includegraphics[width=.5\linewidth]{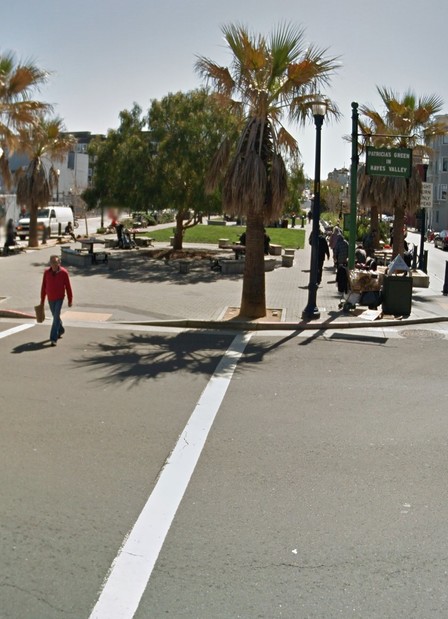}
}\\
\vspace{-.125in}
\subfloat[Crops of the five input panoramas.]{
  \centering
  \includegraphics[width=.2\linewidth]{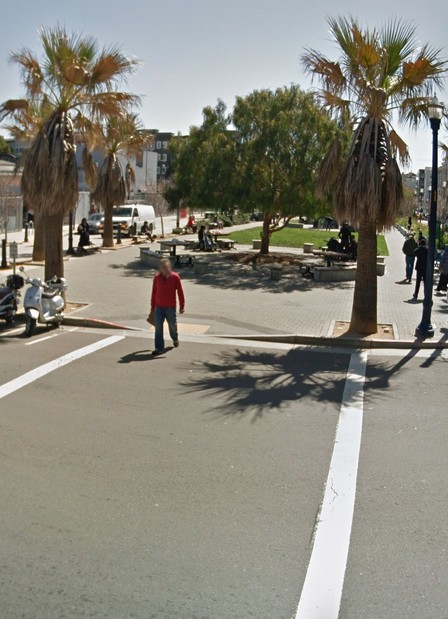}
  \includegraphics[width=.2\linewidth]{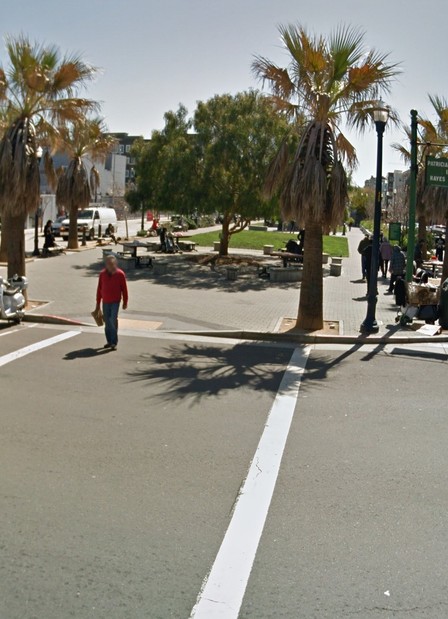}
  \includegraphics[width=.2\linewidth]{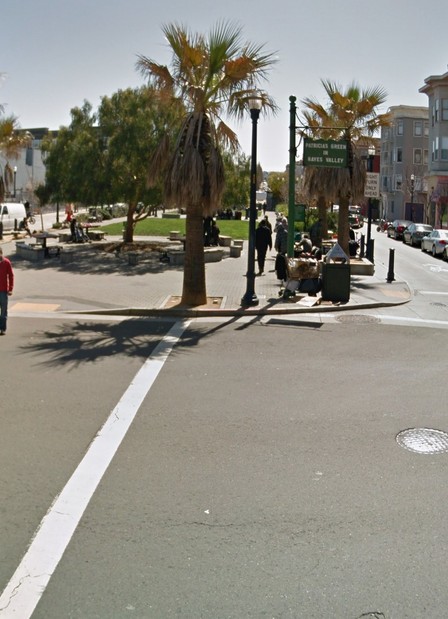}
  \includegraphics[width=.2\linewidth]{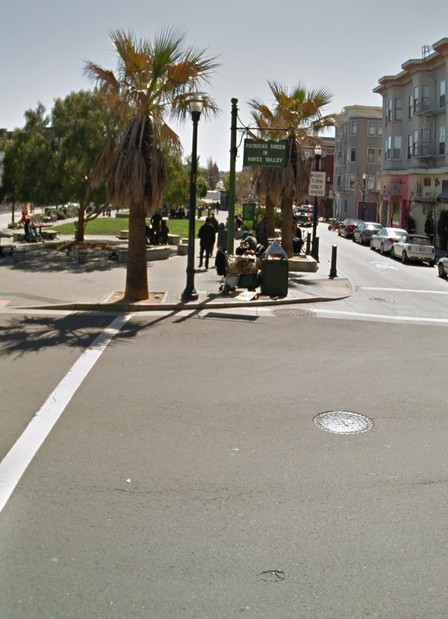}
  \includegraphics[width=.2\linewidth]{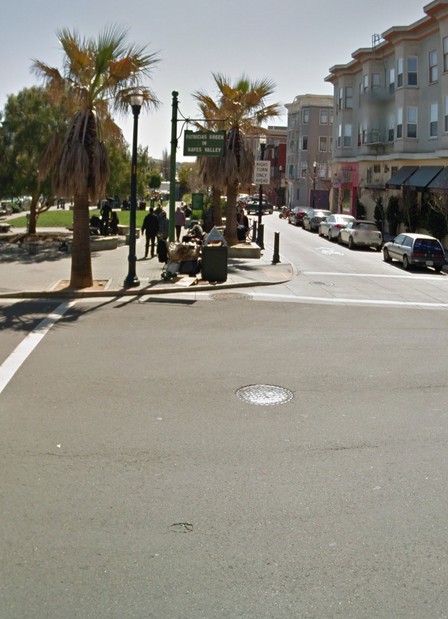}
}
\vspace{-.125in}
\caption{San Francisco park.}
\label{fig:results_sf}
\end{figure}

\begin{figure}%
\centering
\subfloat[Our result.]{
  \includegraphics[width=.5\linewidth]{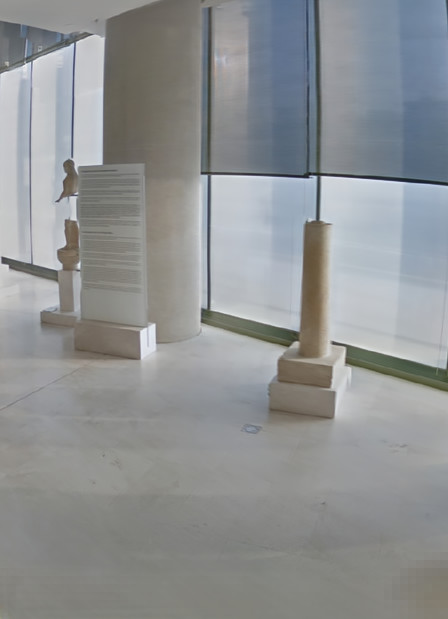}
}
\subfloat[Reference image.]{
  \includegraphics[width=.5\linewidth]{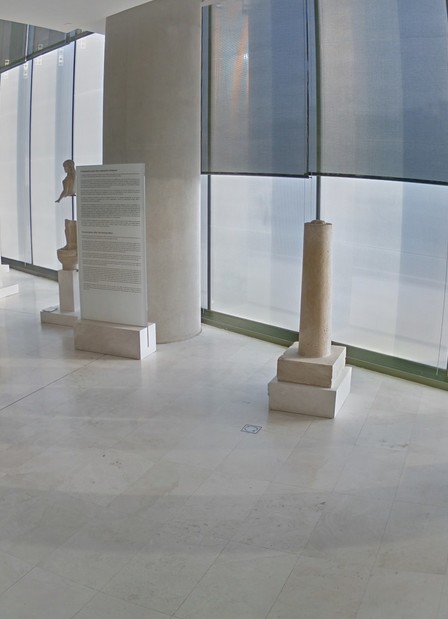}
}\\
\vspace{-.125in}
\subfloat[Crops of the five input panoramas.]{
  \centering
  \includegraphics[width=.2\linewidth]{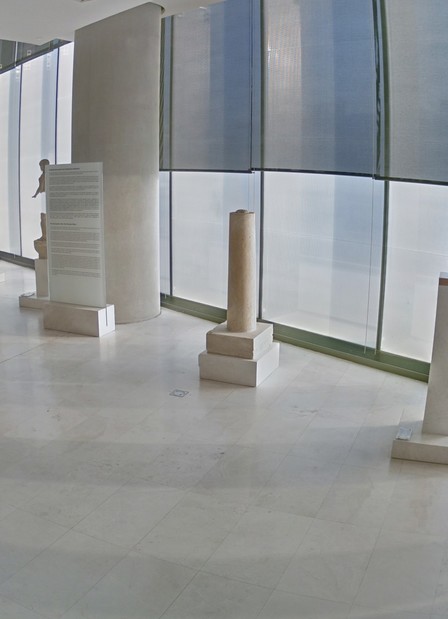}
  \includegraphics[width=.2\linewidth]{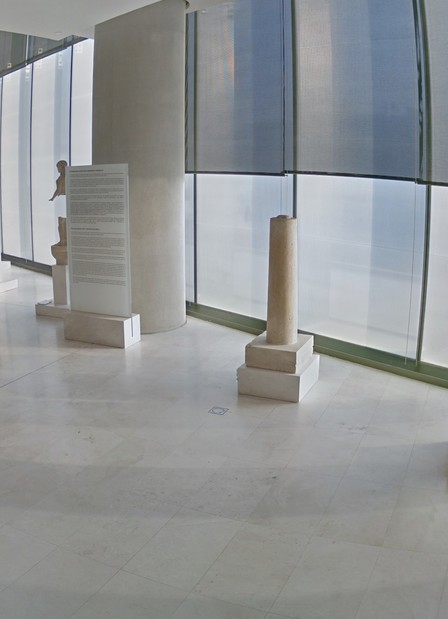}
  \includegraphics[width=.2\linewidth]{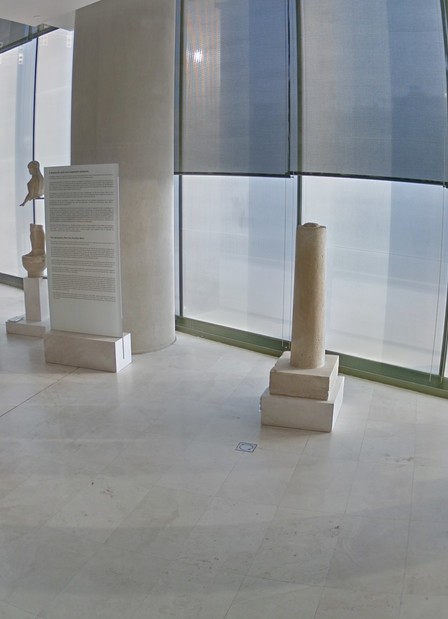}
  \includegraphics[width=.2\linewidth]{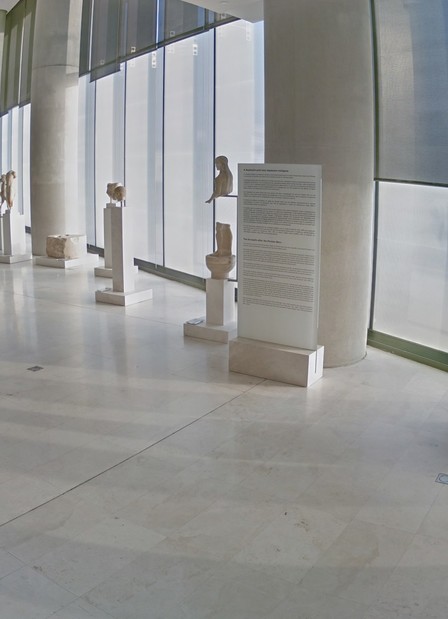}
  \includegraphics[width=.2\linewidth]{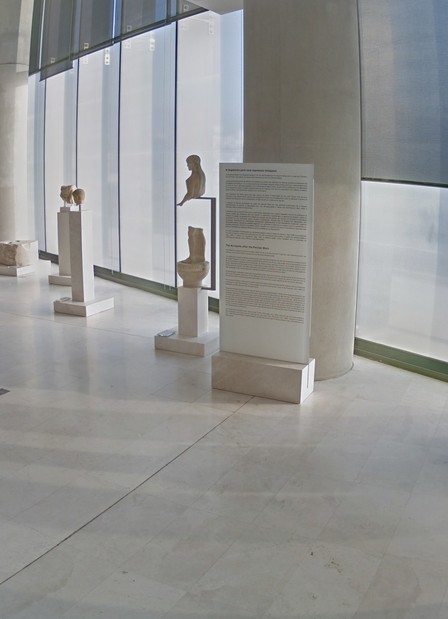}
}
\vspace{-.125in}
\caption{Acropolis Museum.}
\label{fig:results_acropolis}
\end{figure}

\begin{figure}%
\centering
\subfloat[Our result.]{
  \includegraphics[width=.5\linewidth]{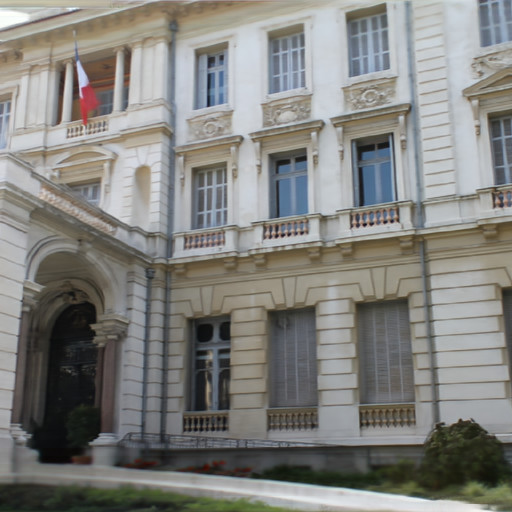}
}
\subfloat[Reference image.]{
  \includegraphics[width=.5\linewidth]{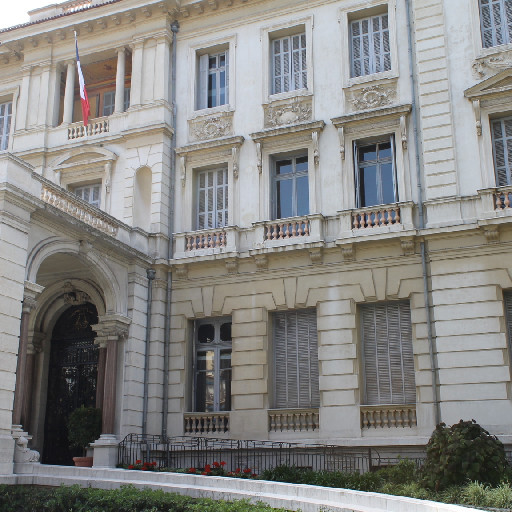}
}
\vspace{-.125in}
\caption{Our method applied to images from \cite{chaurasia13depth}.}
\label{fig:results_mit}
\end{figure}

To evaluate our model on view interpolation, we generated a novel image from the
same viewpoint as a known (but withheld) image captured by the Street View
camera. Representative results for an outdoor scene are shown in
Figure~\ref{fig:results_sf}, and for an indoor scene in
Figure~\ref{fig:results_acropolis}. We also used our model to interpolate from
image data featured in the work of Chaurasia,
\etal~\cite{chaurasia11silhouette,chaurasia13depth}, as shown in
Figure~\ref{fig:results_mit}. The imagery from this prior work is quite
different from our training images, as these prior images were taken with a
handheld DSLR camera. Despite the fact that our model was not trained directly
for this task, it did a reasonable job at reproducing the input imagery and at
interpolating between them.

These images were rendered in small patches, as rendering an entire image would
be prohibitively expensive in RAM. It takes about 12 minutes on a multi-core 
workstation to render a $512\times512$ pixel image. However, our current 
implementation does not fully exploit the convolutional nature of our model, 
so these times could likely be reduced to minutes or even seconds by a GPU 
implementation in the spirit of Krizhevsky, \etal~\cite{krizhevsky12}.

Overall, our model produces plausible outputs that are difficult to immediately
distinguish from the original imagery. The model can handle a variety of
traditionally difficult surfaces, including trees and glass as shown in
Figure~\ref{fig:front_page_result}. Although the network does not attempt to
model specular surfaces, the results show graceful degradation in their
presence, as shown in Figure~\ref{fig:results_acropolis} as well as the supplemental video. 

As the above figures demonstrate, our model does well at interpolating
Street View data and is competitive on the dataset from
\cite{chaurasia13depth}, even though our method was trained on data which has different 
/characteristics from the imagery and cameras in this prior dataset. 
Noticeable artifacts in our results include a slight loss of resolution and the
disappearance of thin foreground structures. Additionally, partially occluded
objects tend to appear overblurred in the output image. Finally, our model is
unable to render surfaces that appear in none of the inputs.

Moving objects, which occur often in the training data, are handled gracefully
by our model: They appear blurred in a manner that evokes motion blur (e.g. see
pedestrians in Figure~\ref{fig:results_sf}).  On the other hand, violating the
maximum camera motion assumption significantly degrades the quality of the
interpolated results. 

\section{Discussion}\label{sec:discussion}
We have shown that it is possible to train a deep network end-to-end to perform
novel view synthesis. Our method is general and requires only sets of posed
imagery. Results comparing real views with synthesized views show the generality
of our method. Our results are competitive with existing image-based rendering
methods, even though our training data is considerably different than the test
sets.

The two main drawbacks of our method are speed and inflexibility in the number
of input images. We have not optimized our network for execution time but even
with optimizations it is likely that the current network is far from real time.
Our method currently requires reprojecting each input image to a set of depth planes; 
we currently use 96 depth planes, which limits the resolution of the output images 
that we can produce. Increasing the resolution would require a larger number of depth 
planes, which would mean that the network takes longer to train, uses more RAM and 
takes longer to run.
This is a drawback shared with other volumetric stereo methods; however, our
method requires reprojected images per rendered frame, rather than just once
when creating the scene. We plan to explore pre-computing parts of the network
and warping to new views before running the final layers.

Another interesting direction of future work is to explore different network
architectures. For instance, one could use recurrent networks to process the
reprojected depth images one depth at a time. A recurrent network would not have
connections across depth, and so would likely be faster to run inference on.  We
believe that with some of these improvements our method has the potential to
offer real-time performance on a GPU.

Our network is trained using 5 input views per target view. We currently can't 
change the number of input views after training which is non-optimal when 
there are denser sets of cameras that can be exploited, as in the sequences from
\cite{chaurasia13depth}. One idea is to choose the set of input views per pixel;
however, this risks introducing discontinuties at transitions between chosen
views. Alternatively, it is possible that a more complex recurrent model could
handle arbitrary numbers of input views, though this would likely complicate
training. It would also be interesting to explore the outputs of the intermediate 
network layers of the network. For instance, it is likely that the network 
learns a strong pixel similarity measure in the select tower. These could be incorporated 
into a more traditional stereo framework.

Finally, a similar network could likely be applied to the problem of
synthesizing intermediate frames in video, as well as for regressing to a depth
map, given appropriate training data.

{\small
\bibliographystyle{ieee}
\bibliography{deepstereo}
}
\end{document}